**EOC2024-Team 7**

## OPTIMIZING THE FLIGHT PATH FOR A SCOUTING UNCREWED AERIAL VEHICLE (UAV)

**Raghav Adhikari**
Department of Mechanical
Engineering
Address
adhika3@clemson.edu

**Sachet Khatiwada**
Department of Civil and Environmental
Engineering
Address
sachetk@g.clemson.edu

**Suman Poudel**
Department of Mechanical
Engineering
Address
spoudel@clemson.edu

**Abstract**

Post-disaster situations pose unique navigation challenges. One of those challenges is the unstructured nature of the environment, which makes it hard to layout paths for rescue vehicles. We propose the use of Uncrewed Aerial Vehicle (UAV) in such scenario to perform reconnaissance across the environment. To accomplish this, we propose an optimization-based approach to plan a path for the UAV at optimal height where the sensors of the UAV can cover the most area and collect data with minimum uncertainty.

## 1. INTRODUCTION

The use of uncrewed vehicles has recently become an emerging area of research. Hu et al. [1] suggested using uncrewed vehicles in civil infrastructure asset management. Similarly, Bechtsis et al. [2] propose using uncrewed ground vehicles (UGVs) in precision farming. One of the emerging areas where such vehicles can prove helpful is assisting in post-disaster evacuation. Natural disasters, including earthquakes, tsunamis, hurricanes, and volcanic eruptions, can severely damage the urban infrastructure, leading to considerable losses. Following such events, providing timely relief and disseminating crucial information, such as safe evacuation routes, becomes essential for affected individuals' safe and organized movement. Recently, among the advanced technologies integrated into disaster response missions include uncrewed aerial vehicles (UAVs) that have been crucial in assessing the state of critical infrastructure essential services, including telecommunications, transportation, and buildings, to facilitate efficient disaster response and evacuation [3].

UAV systems have proven to be increasingly valuable in disaster relief and emergency response (DRER) efforts by enhancing the capabilities of the first responders, offering advanced predictive insights, and enabling early warning systems [4]. UAVs have assisted in diverse tasks, including remote sensing, search and rescue, forest fire detection, survey and surveillance [5]. Using uncrewed vehicles in such scenarios can help minimize the human element in such endeavors and minimize the risk to human lives. In addition, such scenarios often involve compromised visibility, and the presence of sensors that do not rely on vision (e.g., LiDAR) can help overcome these hurdles. When ground-based communication is limited or unavailable in disaster scenarios, UAVs can serve as communication relays, extending the communication range and establishing network connections between dispersed ground units [5].

Equipped with onboard communication devices, UAVS helps improve network connectivity, serving as relay points and extending coverage in remote or post-disaster areas where communication infrastructure is sparse or lacking [6][7]. In these environments, UAVs autonomously gather environmental data through sensors and cameras to assess disaster conditions in real time [8][9]. For instance, after Japan's 2011 Tohoku earthquake and tsunami, the Global Hawk UAV was deployed to monitor the radiation-affected Fukushima Daiichi Nuclear Power Plant [10][11]. These operations illustrate how UAVs can provide critical information on hazardous areas, allowing response teams to evaluate the immediate needs of the affected regions.

The primary goal of using UAVs in DRER missions, especially in situations like tornadoes, is to save lives by providing early warnings and improving response time [12]. UAVs also relieve humans from dangerous tasks, and their deployment is increasingly common in disaster management operations globally. UAVs contribute significantly to subsequent stages of disaster management and the early phases emphasizing search and rescue, including debris removal and infrastructure reconstruction [13].

Furthermore, various companies are now exploring UAV technology that can facilitate the efficient delivery of goods to isolated areas after a disaster. Specialized UAVs are designed that can autonomously deliver refugees food, water, and medical supplies after a disaster [14][15]. UAVs also offer a cost-efficient alternative to satellite images, which can be costly and challenging to interpret. In cases like the Pakistan earthquake, collaborations between institutions provided real-time images to assist relief agencies. UAVs, particularly micro-UAVs, can deliver up-to-date images and even track civilian evacuation routes, offering significant benefits for relief operations [16].

The path planning problem is a common problem, and there are multiple approaches in the literature to address it. One of the earliest approaches was proposed by Dijkstra [17], called the Dijkstra's algorithm, which could find the shortest path between a start point and a target point in a graph by minimizing the cost to travel from the start node to any other node in the graph, until we approach the target node. Soon after, Hart et al. [18] realized that the shortest path problem could be immensely accelerated by using an estimate of the cost to go to the target from each node, famously called the heuristic cost, and proposed the A* algorithm. While these algorithms successfully generated the shortest path for known environments, there was also a need to plan paths for environments where not much was known initially. So, Stentz [19] proposed the D* algorithm, which was able to plan an initial path in a similar fashion to that of the A* algorithm, but it also had the replanning capability to adjust or replan when new information about the environment was known. Koenig et al. [20] built on the D* algorithm to develop the D* lite algorithm, which was faster than the D* algorithm. These algorithms were great for small environments, but for large environments, graph-based algorithms are often found lacking because they take a long time to generate the solution. To remedy that, LaVelle [21] proposed the rapidly exploring random trees (RRT) algorithm, and Karaman et al. [22] further improved upon it to propose the RRT* algorithm. RRT and RRT* are sampling-based algorithms that randomly sample points from across the environment and connect them to form trees that connect the start and the target points. RRT* has an additional step of tree adjustment, which guarantees asymptotic optimality.

These algorithms do well for cases where only one cost is considered. However, the path planning problem encountered in the real world often has multiple costs, and the path planning algorithm needs to be able to find some kind of trade-off between those costs. Cai et al. [23] proposed a multi-criteria path planner, where the different costs were the localization uncertainty of the vehicle, collision risk to the obstacles in the environment, and the distance to the target. Similarly, other works like those by Kurzer [24] and Saikh & Goodrich [25] also propose path planners that incorporate multiple cost components. Shen et al. [26] and Yu et al. [27] proposed path planners that considered multi-criteria costs, which consist of vehicle limitations and the smoothness of the path generated. Braun [28] proposed another approach to quantify the different aspects of cost by using a single value, which was the real cost to move the vehicle across some links in the environment. This cost was then used to estimate the cost of the untraveled links in the environment.

Optimization algorithms are used to find numerical solutions for optimization problems. It involves maximizing or minimizing an objective function. Convex programming minimizes an objective function, while concave programming maximizes an objective function. Quadratic programming is a form of convex programming where an objective function includes a quadratic term. This can be applied to path planning problems in autonomous driving systems (ADS) to minimize path length.

Viana et al. [29] proposed a Distributed Model Predictive Control (MPC) incorporating the Human Driver Model (HDM) with Mixed-integer quadratic programming (MIQP), named the HDM-MIQP algorithm. This algorithm includes the HDM into the MPC model, assuming HDM errors and uncertainties as constant values. Similarly, Kanchwala applied the HDM-MIQP algorithm under the same project of the bicycle model with different vehicle dynamics. He formulated the model using the CarSim driving simulator and assuming fixed HDM uncertainties and errors.

Oliveira et al. [30] applied Sequential Quadratic Programming (SQP) for motion planning, using a road-aligned car model and a new approximation method. It showed better performance than previous algorithms but without considering approximation error. Zhu et al. proposed a Parameterized Curvature Control (PCC) algorithm where the cubic spline interpolation technique is used as a parameterization method. The cubic spline is better than a single polynomial because the cubic spline can represent all possible curves with curvature. Then, the splined optimization problem is solved by the SQP algorithm. The results proved that the PCC algorithm obtained suitable safe paths in a dynamic environment.

Changhao et al. [31] developed a Dynamic Programming-based algorithm integrated with Clothoid Curve (DPCC) algorithm. The concept is to obtain future $T$ samples and then fit them into a clothoid curve path by solving it as a dynamic programming problem. This algorithm found an acceptable path with acceptable performance. However, more scenarios need to be tested. Moreover, the parameters of the algorithm were not optimized. Furthermore, the algorithm is not compared to any other algorithm.

Hu et al. [32] proposed an algorithm for road intersection driving scenarios called the Event-Triggered Model Predictive Adaptive Dynamic Programming (ET-MPADP). The ET-MPADP algorithm integrates the MPC controller's control policy generation and the mismatch of cost function approximation using the critic actor scheme. The ET-MPADP algorithm was better than the time-triggered MPADP (TT-MPADP) algorithm. However, the structure of regression needs to be considered. Moreover, the unknown motion of the obstacles is not considered as well.

Jiang et al. [33] introduced the QP-SQP algorithm, a quadratic programming-based path planner for static, cluttered environments. This algorithm has two stages: first, it generates a collision-free guideline using Quadratic Programming (QP), and second, it produces an optimal path using the Sequential Quadratic Programming (SQP) technique, with both optimization problems solved by the Open-Source Quadratic Programming (OSQP) solver. Simulation results suggest that the QP-SQP algorithm outperforms the Piecewise Jerk Method (PJM) and Minimum Snap Method (MSM), but further testing with deep statistical analysis is needed.

Optimization-based methods are generally time-consuming that requires high computation times to solve the problem. And also, these methods are problem dependent. An optimizer that works well in one driving scenario may fail in another. Future directions for utilizing these methods include hybridization with other techniques. Hybrid algorithms combine sampling-based methods with quadratic programming, such as the SB-SQB and S-NO algorithms [34]. These techniques reduce computational complexity while improving real-world adaptability for more efficient and practical solutions.

## 2. PRELIMINARIES

We have an environment that is represented by a 3-dimensional occupancy grid. The occupancy grid represents the parts of the environment that are occupied by obstacles (e.g., buildings, trees, etc.). The UAV must move from a start $S \in \mathbb{R}^3$ to a target $T \in \mathbb{R}^3$. The UAV starts from the start position $S$. The position of the UAV at time '$t$' can be represented by $p_t = [x_t, y_t, z_t]$. We are not dealing with its rotation because we assume a multirotor UAV, and we assume that the positions between any timesteps $t$ and $(t + 1)$ can be handled by the dynamics of the UAV.

## 3. METHODOLOGY

The environment in question is an occupancy grid that contains probability that each cell is occupied. Values close to 1 represent a high probability that the cell contains an obstacle, and values close to 0 represent a high probability that the cell is a vacant space. Since the occupancy map is 3-dimensional, it is computationally expensive to run an optimization-based path planner that must find 3*n decision variables, where 'n' is the number of waypoints along the path. So, to simplify the problem, we tried to break this complex optimization problem into two simpler parts:

a) Part 1: 2D path planning
b) Part 2: Maintaining the scouting height

### 3.1 Part 1: 2D Path Planning

The first part involves planning the waypoints for the UAV in a 2D plane. We are working under the following assumptions:

a) We have an environment that has obstacles which do not have increasing cross sections as we move up the Z-axis (e.g., trees)
b) Obstacles that have increasing cross sections as we move up the Z-axis have been represented by a prism of the cross section equal to their maximum cross section

The above assumptions allow us to plan a path for the UAV on a 2D XY-plane, and later adjust the flight coordinates to optimal scouting height. So, to accomplish the first task, we projected the 2D occupancy map onto the XY-plane. This represents the navigable space for the UAV, disregarding the vertical information. In the planning process, the UAV's current position is defined as its starting position and a target position is explicitly defined. Since we are looking for waypoints, the path will be planned for a point object. So, to avoid collision checks, we inflated the obstacles in the 2D map to accommodate for the UAV's physical dimensions. This ensures the generated paths avoid collisions with the nearby objects. Figure 1 shows a randomly generated 2D occupancy grid before and after inflating the obstacles.

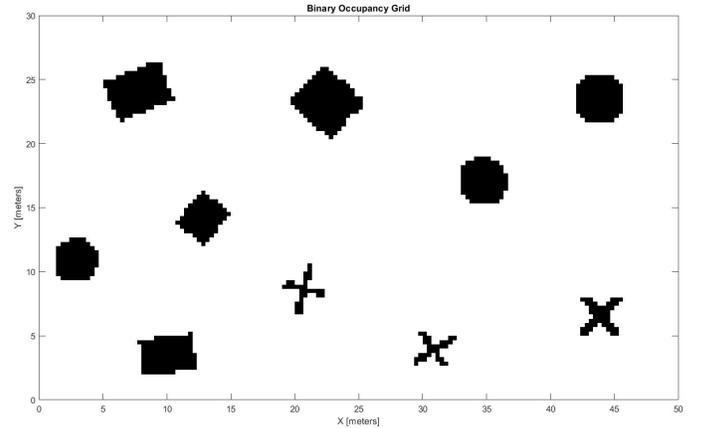

*Figure 1: Randomly generated 2D occupancy grid*

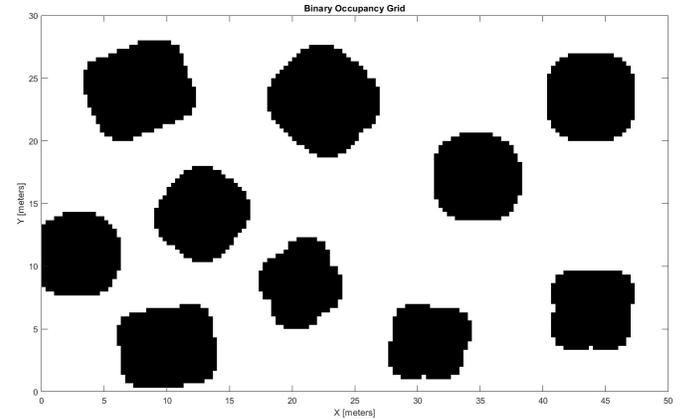

*Figure 2: 2D occupancy grid with inflated obstacles*

Since finding out the whole path at a time can be computationally challenging and time consuming, we formulated the optimization process such that it attempts to find the next best step, i.e. we plan one step at a time. The mathematical formulation for the first part of the problem was defined as follows:

### 3.1.1 Objective

We have two objectives: first one is to get to the target point and second one is to avoid the obstacles while doing so. To incorporate these two objectives, we use a weighted multi-objective function as:

$$minimize Z_a = w_1 * \left\| p_t - T \right\|_2^2 + w_2 * Occupancy(p_t)$$

Where $p_t$ = UAV position in XY plane at timestep t
$w_1, w_2$ = weights assigned to each objective
$Occupancy(p_t)$ = Occupancy value at point $p_t$

### 3.1.2 Constraints

The constraints on part 1 of the problem were formulated as follows:

a) Constraints on the step size: Since we wanted to avoid erratic UAV movements, we had to constraint the maximum displacement the UAV can take between successive timesteps. Thus, we introduced a constraint on the step size as:

$$C_{a1}: \left| \left| p_t - p_{t-1} \right| \right|_2 \leq c_s$$

Where $p_t$, $p_{t-1}$ = UAV position in XY plane at timestep t and t-1 respectively

$c_s$ = Allowed maximum step size

b) Constraints on the starting UAV position at each optimization step: At each timestep 't', the starting location of the UAV had to be fixed to the location obtained from the previous optimization, i.e. $p_{t-1}$.

$$C_{a2}: p_s = p_{t-1}$$

Where, $p_s$ = The starting position of the UAV at each timestep ($p_s = S$, at t = 1)

c) Constraints on the environment boundary: We also did not want the path to drift beyond the environment boundary. Thus, we introduced boundary constraints on the decision variables which place them within the environment boundary.

We carried out the optimization recursively, finding the next best step to go to at each timestep. At each timestep (t) the decision variable was the position ($p_t$). This generated a set of waypoints for the UAV, on the XY plane, from the starting location to the target location. These waypoints were then used in the next step to optimize the flight altitude.

### 3.2 Part 2: Maintaining Scouting Height

We generated waypoints in the XY plane for a collision-free optimal path in the first part. Since the obstacles have already been avoided, the goal of the second part of the process is to determine the optimal flight heights along those waypoints. We assume that there is a certain height 'h' for a given UAV that is optimal for data collection. Flight beyond the optimal height will hamper the quality of the data, and below that height will reduce the area for which the data is collected.

The mathematical formulation for the second part was done as follows:

### 3.2.1 Objective

The objective for this part of the problem is to maintain the optimal height, which was formulated as:

$$minimize Z_b = \left| \left| Z - h \right| \right|_2^2$$

Where $Z \in \mathbb{R}^n$
n = number of waypoints along the path
h = Optimal data collection height for the UAV

### 3.2.2 Constraints

The constraints on part 1 of the problem were formulated as follows:

a) Constraints on the step size: We wanted to avoid erratic UAV jumps. Therefore, we introduced a constraint on the change in height between successive UAV positions as:

$$C_{b1}: z_{t+1} - z_t \leq c_z$$

$$C_{b2}: z_{t+1} - z_t \geq -c_z$$

Where $z_t, z_{t+1} \subset Z$, are UAV heights at timesteps t and (t+1) respectively

$c_z$ = Maximum allowable change in height between successive timesteps

b) Constraints on the start and target height: We assume that the UAV starts and ends at known locations with known height.

c) Boundary constraints: The possible flight heights were constrained between a minimum of zero to a maximum of the maximum possible flight height the drone is capable of.

The decision variable was the height vector (Z) of size n = the number of waypoints generated, which represents the UAV height at each waypoint/timestep. This would complete the 3D trajectory of the scouting UAV from the starting location to the target location.

## 4. RESULTS

We used the "fmincon" solver, available in MATLAB, with Sequential Quadratic Programming (SQP), to solve both of our optimization problems. We first tested the 2D path planner on a randomly generated 2D occupancy grid. Then we tested it using the "dMapCityBlock" 3D occupancy grid available in MATLAB's "Navigation Toolbox" as our planning environment. For the 3D occupancy map, we first projected it onto the XY plane to generate a 2D occupancy map, before planning the path.

For part 1, we used the maximum allowable step size ($c_s$) of $\sqrt{2}$. We tuned the weights ($w_1 \& w_2$) based on the map used. We kept the weight $w_2$ constant but changed $w_1$ based on the current position of the UAV such that it increases as the search gets closer to the target location. For the second part, we assumed the optimal scouting height of 35 meters, and the absolute change in height between successive timesteps/waypoints ($c_z$) to 1 meter.

### 4.1 Planning on randomly generated occupancy grid

We randomly generated a 2D occupancy map for testing the first part of our approach. Figure 3 shows the map generated, where the occupied cells (obstacles) are represented in black and free cells are represented in white. The path planned on the map is shown in figure 4.

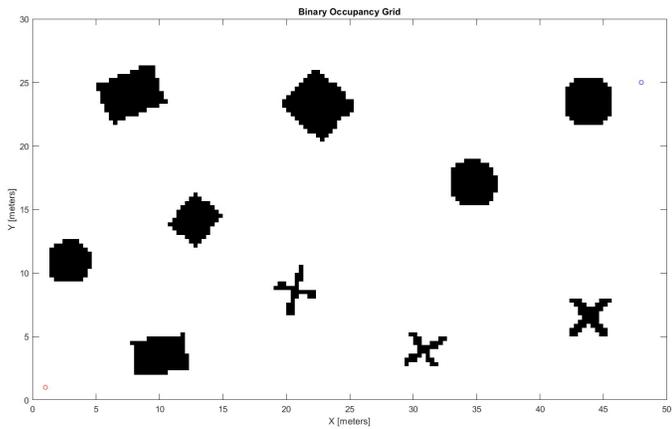

*Figure 3: Randomly generated 2D obstacle grid*

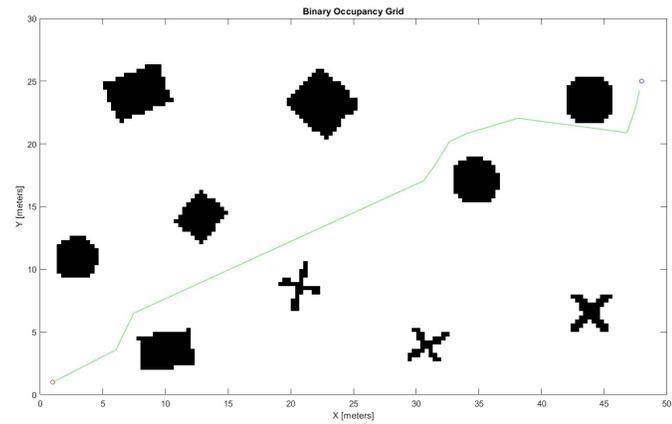

*Figure 4: Path 2D planned*

### 4.2 Planning on 3D occupancy map

We used the "dMapCityBlock" 3D occupancy grid available on the "Navigation toolbox" in MATLAB shown in figure 5 to test our approach. Figure 6 shows the 2D projection of the environment. The path generated on the 2D XY plane is shown in figure 7. The final trajectory is shown in figure 8.

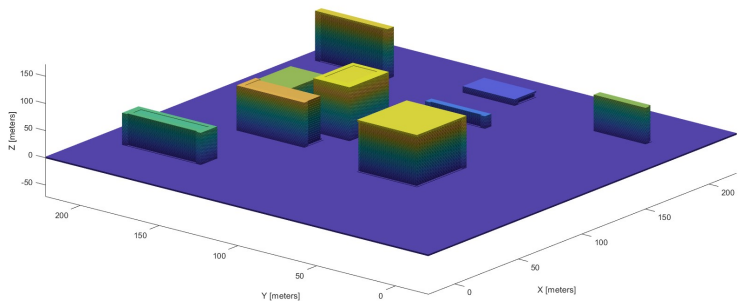

*Figure 5:MATLAB's "dMapCityBlock" 3D occupancy grid*

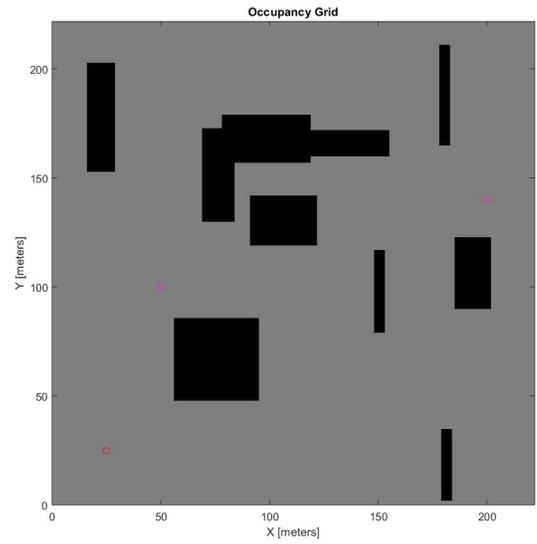

*Figure 6: 2D projection of the 3D occupancy grid on the XY plane (Red circle shows the start location, and the magenta circle shows the target location)*

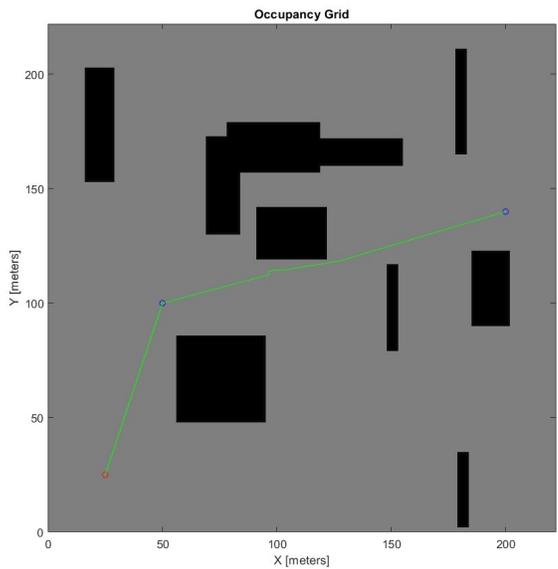

*Figure 7: Planned 2D path on the projected grid*

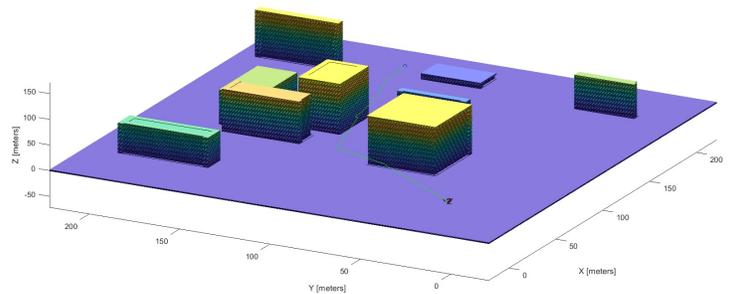

*Figure 8: Planned 3D trajectory for the UAV*

     

## DISCUSSION AND CONCLUSION

We were successfully able to generate trajectories for a scouting UAV. Since we used a step-by-step optimization approach, we were able to see the progression of the path. Some of the key discussion points we observed are:

a) The optimization progressed smoothly along the areas that were not near the obstacles. However, for areas that are closer to obstacles significantly slowed down the optimization process.

b) Tuning the weights was a challenge, because the scale of the two objectives were different so it was hard to find weights that would balance the two objectives.

c) For large environments, like the 3D occupancy grid used, just using the dynamic weights were not enough. The path would often drift into the obstacles. So, we decided to use an intermediate point to balance the dynamic weighting.

d) We could have used the obstacles as constraints, but that significantly slowed down the optimization without good enough results.

Some of the limitations of our approach are:

a) As mentioned before, the tuning for the weights was challenging. This can potentially be handled using an additional optimization loop that manages the weights if the obstacles are approached.

b) Since we are assuming all obstacles are prismatic, and the non-prismatic obstacles have been represented by prisms with cross sections representing the maximum cross section of such obstacles, we are sacrificing significant free space. This can potentially be avoided with better modelling for the obstacles, which could then make the problem more challenging.

Thus, the flight path for a scouting UAV was accomplished in a 3D environment by breaking down the problem into 2D path generation problem and then maintaining the optimal scouting height. We successfully generated paths for 3D environments where the obstacles were generated by an occupancy grid. Further works can investigate improvement opportunities in tuning weights, better representations of obstacles, and better obstacle avoidance.